\documentclass[10pt, conference, compsocconf]{IEEEtran}
\ifCLASSINFOpdf
\else
\fi

\usepackage{amsmath}
\usepackage{mathtools}
\usepackage{graphicx}
\usepackage{float}
\usepackage{adjustbox,lipsum}
\usepackage{amssymb}
\usepackage{algorithmic}
\usepackage{algorithm}
\usepackage{caption}

\DeclareCaptionLabelFormat{noname}{#2}
\newcommand\mat[1]{\mathcal{#1}}
\begin{document}
\title{Ranky : An Approach to Solve Distributed SVD on Large Sparse Matrices}

\author{\IEEEauthorblockN{Resul Tugay}
\IEEEauthorblockA{Department of Computer Engineering\\
Istanbul Technical University\\
Istanbul, Turkey\\
Email: tugayr@itu.edu.tr}
\and
\IEEEauthorblockN{ \c{S}ule G\"{u}nd\"{u}z \"{O}\c{g}\"{u}d\"{u}c\"{u}}
\IEEEauthorblockA{Department of Computer Engineering\\
Istanbul Technical University\\
Istanbul, Turkey\\
Email: sgunduz@itu.edu.tr}
}

%Telephone: (800) 555--1212, Fax: (888) 555--1212}


\maketitle

\begin{abstract}
Singular Value Decomposition (SVD) is a well studied research topic in many fields and applications from data mining to image processing. Data arising from these applications can be represented as a matrix where it is large and sparse. Most existing algorithms are used to calculate singular values, left and right singular vectors of a large-dense matrix but not large and sparse matrix. Even if they can find SVD of a large matrix, calculation of large-dense matrix has high time complexity due to sequential algorithms. Distributed approaches are proposed for computing SVD of large matrices. However, rank of the matrix is still being a problem when solving SVD with these distributed algorithms.  In this paper we propose Ranky, set of methods to solve rank problem on large and sparse matrices in a distributed manner. Experimental results show that the Ranky approach recovers singular values, singular left and right vectors of a given large and sparse matrix with negligible error.

\end{abstract}


% \fi
\IEEEpeerreviewmaketitle

\section{Introduction}
\label{sec:Intro}
The Singular Value Decomposition (SVD) of a matrix $\mat{A}$ is the factorization or decomposition of the matrix into the product of three matrices. Formally, the singular value decomposition of a matrix $\mat{A}$ with $M$ rows and $N$ columns  can be represented as
$\mat{U} \Sigma \mat{V}^* $
where $\mat{U}$ is a unitary matrix $(\mat{U}^T = \mat{U}^{-1})$ with dimensions $M$x$M$, $\mat{V}^*$ (conjugate transpose of $\mat{V}$) is also a unitary matrix having $N$x$N$ dimension and $\Sigma$ is $M$x$N$ diagonal matrix with non-negative real diagonal numbers where $\Sigma_{ii} = \sigma_i$ for $i = 1,...,min(M,N)$. If the matrix $\mat{A}$ is real, then $\mat{U}$ and $\mat{V}$ are real and orthogonal. The vectors $u_i$ $(i=1,...,M)$ and $v_j$ $(j=1,...,N)$ are called the left and right singular vectors respectively and $\sigma_k$ $(k=1,...,min(M,N))$ are the singular vectors. In this paper it is assumed that the matrix $\mat{A}$ is 'short and fat' where the number of columns are much more than the number of rows. But the matrix can also be 'tall and skinny' matrices where row numbers are much more than number of columns.

It is possible to get singular components of $\mat{A}$ by finding eigenvalues and eigenvectors of cross product matrices ($\mat{A}^* \mat{A}$ and $\mat{A} \mat{A}^*$). The left and right singular vectors are the eigenvectors of the matrices and singular values are the nonnegative square roots of the eigenvalues of one of the cross product matrices. Besides, singular components can be found by finding eigenvalues and eigenvectors of a symmetric matrix called cyclic matrix that is constructed as a matrix ($[\mat{A}|\mat{A}^*]$) from $\mat{A}$ and $\mat{A}^*$. But these two approaches are not recommended to compute singular components because of high cost of the computation of cyclic matrix especially the matrix is sparse. Additionally, these methods cause loss of accuracy when computing $\mat{A} \mat{A}^*$ \cite{iwen2016distributed}.
Generally SVD algorithms focused on bidiagonalization step in order to get cross product matrix without computing it explicitly \cite{cline2006computation}. Householder method is one approach for computation of the bidiagonal form of a given matrix $\mat{A}$ and Golub-Kahan bidiagonalization or Lanczos bidiagonalization \cite{golub1965calculating} is another approach.

On top of these core algorithms, there are several distributed algorithms which can run simultaneously. The complexity of computing the SVD is $O(M^2N)$ or $O(MN^2)$ where $M<N$ or $M>N$ respectively for a matrix of size $M$x$N$. These algorithms try to solve the SVD problem with less complexity by using distributed incremental or hierarchical algorithms.
Recently, Iwen and Ong \cite{iwen2016distributed} proposed an algorithm to construct SVD of a matrix in a distributed and incremental way. 
The algorithm is able to recover singular components of large, dense and highly rectangular matrices, but not sparse matrices. SVD of a matrix can only be solved if its rank is known according to their algorithm. When the dimension of a matrix with the size of $M$x$N$ is considered, rank of the matrix will be at most min($M,N$), which is number of rows in their assumption.
%SGO: burda bir problem var gibi. rank en fazla min(M,N) olabilir diyorsun, yani daha kucuk de olabilir. asagida da rank < min(M,N) olursa problem var diyorsun. rank nasil bir problem oluyor? literaturde tall thin matrix icin SVD yontemleri var (Fast low-rank modifications of the thin SVD)

%hocam düşük rank distributed yaptığımız için problem oluyor, propose edilen algoritma ancak rank eşitken çalışıyor.
%SGO: o zaman sana gönderdiğim yöntemlere referans vererek onların distributed olarak çalıştırılamayacağını (bu doğruysa) söylemen gerekiyor.
%R : Tamam hocam üst cümlede belirttim
If the input matrix is large and sparse, rank of the block matrices which are parts of the input matrix will be smaller than min($M,N$). This situation of rank or unknown rank causes undetermined results when computing the SVD. We propose Ranky, set of methods, called \textit{RandomChecker}, \textit{NeigborChecker} and \textit{NeigborRandomChecker} aim to solve the rank problem for large and sparse matrices.
%SGO: burda bahsettiğin block matrix ikinci paragrafta bahsettiğin block matrix ni, yoksa related work'te bahsettiğin block matrix'mi? eğer related work'tekiyse burda bu şekilde söyleyince anlamsız oluyor.
%R : İkinci paragraftaki block matrisi kavramını kaldırdım hocam kafa karışmasın diye, diğer tüm block matrisler aynı anlamda, matrisin parçalanmış hali.
\iffalse
To summarize we make the following contributions:
\begin{itemize}  
\item We propose three methods to solve the rank problem of large and sparse matrix. 
\end{itemize}
\fi

The rest of the paper is structured as follows: Section \ref{sec:RW} presents related work and section \ref{sec:BG} gives details about proposed methodology. Experimental results and analysis are shown in section \ref{sec:ER}. Finally, section \ref{sec:FW} concludes the paper.

\section{Related Work}
\label{sec:RW}

SVD  has many useful applications in many fields from data mining to signal processing including PCA \cite{hannachi2007empirical} and data clustering \cite{dhillon2001co}% , dimension reduction \cite{ravi1998dimensionality} and image restoration \cite{wall2003singular} \cite{moonen1995svd}. 
Although the history of the SVD dates back to 1900, it was first established for general rectangular matrices by Eckart and Young \cite{eckart1939principal} in 1939. Then it has become more and more popular after that year. 

The SVD of a matrix can be formulated as an eigenvalue problem. Compared with an eigenvalue problem, it only works on some of square matrices, but SVD can be applied to all types of (square,rectangular) matrices. Input matrix must be transformed to square matrix before the SVD problem is considered as eigenvalue problem. There are two possible ways to achieve this:
\begin{itemize}  
\item The cross product matrix, either $\mat{A}^* \mat{A}$ or $\mat{A} \mat{A}^*$
\item The cyclic matrix
$
H(\mat{A}) = 
\begin{bmatrix}
0 & \mat{A} \\
\mat{A}^* & 0 
\end{bmatrix}
$
\end{itemize}

  Roman \textit{et al.} stated that these two approaches are not feasible to get singular components of non-square input matrix due to their drawbacks in chapter 4 in \cite{roman2015slepc}. Then Golub and Kahan \cite{golub1965calculating} proposed bidiagonalization algorithm to solve the SVD problem. This algorithm produces the partial bidiagonal reduction of input matrix with increasing dimension in each iteration. There are several implementation of this algorithm in literature. Golub \textit{et al.} \cite{golub1981block} implemented block version of this method. A good low rank approximation algorithm, a version of lanczos bidiagonalization called one-sided SVD, was proposed by Simon and Zha \cite{simon2000low}. Once bidiagonal form of input matrix is calculated, then the singular values of bidiagonal matrix form can be calculated using $QR$ algorithm \cite{golub1970singular}. There is also stable divide and conquer algorithm proposed by Gu and Eisenstat \cite{gu1995divide} to compute the SVD of lower bidiagonal matrix.

In some big data applications, the input data is represented as a short and fat matrix with a small number of samples having a large set of features or vice-versa. For instance, there are only ten thousands of terms in Wikipedia, while the number of articles has more than 5.5 millions. 
There are several studies in literature attempted to solve distribute SVD for non square matrices \cite{qu2002principal,bai2005principal,baker2012low,menon2011fast,brand2006fast}.
Qu \textit{et al.} proposed a distributed SVD algorithm for tall and skinny matrices and reported the results on synthetic data. Although they have a good accuracy, their algorithm works efficiently when the local matrices have low ranks \cite{qu2002principal}. Another algorithm proposed in \cite{bai2005principal} that is based on the algorithm proposed in \cite{qu2002principal} is using hierarchical QR algorithm to solve PCA and inherently the SVD problem. This distributed algorithm uses tree-based merge technique to collect and merge R matrices. 
Iwen and Ong proposed an algorithm called a distributed and incremental algorithm for short and fat matrices.
The idea behind this algorithm is computing SVD of block matrices of the input matrix separately, then concatenating singular values and left singular vectors of the block matrices to create a proxy matrix and recovering SVD from the proxy matrix. Dimension of proxy matrix is much more smaller than the original input matrix because of highly rectangular matrix. More recently, Vasudevan and Ramakrishna proposed a hierarchical SVD algorithm for low-rank matrices, matrices were large and dense but inherently low-rank \cite{vasudevan2017hierarchical}. Their algorithm is not working with only short and fat or tall and skinny matrices also all types of other matrices. Further, they split the matrix into blocks, both row-wise and column-wise unlike the Iwen and Ong algorithm. But this algorithm is suitable for the dense and low rank matrices not for large and sparse matrices. Also Edo \cite{liberty2013simple} proposed a deterministic matrix sketching algorithm that provides a sketch matrix $B$ which is a good approximation of A. But as he stated, the algorithm does not consider sparse matrices. Then Ghashami \textit{et al.} \cite{ghashami2016efficient} proposed a variant of the sketching algorithm for sparse matrices. However, their algorithm is aiming to create a compact matrix that is a good assumption of another large matrix. 
%SGO: rectangular demeyelim, fat matrix olabilir
%SGO: hala cogu rectangular diye geciyor. fark etmiyor mu? fat veya tall olsa da ayni yontem uygulanabilir mi?

%evet hocam fat ise column-wise, tall ise row-wise blocklara ayrılarak yapılabiliyor.

%SGO: HALA RECTANGULAR DIYE BAHSEDIYORSUN

%%R: Hocam referans verdiğim paper'da highly rectangular matrix olarak geçtiği için bu kısımları değiştirmedim fakat diğer tüm yerlerde short and fat matrix olarak değiştirdim. 

\section{Background and Methodology}
\label{sec:BG}
In this section, the rank problem will be discussed in detail on large and sparse matrices. Firstly, distributed and incremental SVD algorithm and rank problem then proposed methods will be described to overcome the problem.

Let $\mat{A}$ $\in$ $\mathbb{C}^{MxN}$ be the input matrix and $\mat{A}^i$
$\in$ $\mathbb{C}^{MxN_i}, i = 1,2,...,D$ be the block decomposition of $\mat{A}$ where 
$\mat{A}$ = $[\mat{A}^1|\mat{A}^2|...|\mat{A}^D]$. Since $\mat{A}^i$ has a rank at most $d \in \{1,...,M\}$, each block has a reduced SVD representation,
\begin{equation}
\label{eq1}
\mat{A}^i = \sum_{j=1}^{d} u_j^i \sigma_j^i (v_j^i)^* = \hat{U}^i \hat{\Sigma}^i \hat{V}^{i*} 
\end{equation}

Let $\mat{P}$ = $[ \hat{U}^1 \hat{\Sigma}^1   | \hat{U}^2 \hat{\Sigma}^2 |...| \hat{U}^D \hat{\Sigma}^D ]$ be the proxy matrix of $\mat{A}$. If $\mat{A}$ has the reduced SVD decomposition, $\mat{A} = \hat{U} \hat{\Sigma} \hat{V}^{*}  $ and $\mat{P}$ has the reduced SVD decomposition, $\mat{P} = \hat{U}^\iota \hat{\Sigma}^\iota \hat{V}^{\iota*} $, then $\hat{\Sigma} = \hat{\Sigma}^\iota $, and $ \hat{U} = \hat{U}^\iota W$ where $W$ is a unitary  block diagonal matrix.
As it is stated earlier, the singular values of $\mat{A}$ are the (non-negative) square root of the eigenvalues of $\mat{A}\mat{A}^*$. Then,
\begin{equation}\label{AA}
\begin{split}
\mat{A}\mat{A}^* &= \sum_{i=1}^{D} U^i \Sigma^i (V^i)^* (V^i)((\Sigma^i)^*)(U^i)^* \\
&= \sum_{i=1}^{D} U^i \Sigma^i((\Sigma^i)^*)(U^i)^*
\end{split}
\end{equation}
Similarly, the singular values of $\mat{P}$ are the (non-negative) square root of the eigenvalues of $\mat{P}\mat{P}^*$.
\begin{equation}\label{PP}
\mat{P}\mat{P}^* = \sum_{i=1}^{D} U^i \Sigma^i (U^i \Sigma^i)^* 
= \sum_{i=1}^{D} U^i \Sigma^i((\Sigma^i)^*)(U^i)^*
\end{equation}
%SGO. bu tanimlar icin referans verebilir miyiz?
%R : Evet hocam, İwen and Ong olarak verdim aşağıda.
Iwen and Ong proved equation \ref{AA} and \ref{PP} respectively \cite{iwen2016distributed}. It is concluded that SVD of proxy matrix $\mat{P}$ must be the same as SVD of the matrix $\mat{A}$ if and only if each block ($\mat{A}^i$) of the matrix $\mat{A}$ has rank $d$. The incremental (hierarchical) SVD algorithm proposed by Iwen and Ong \cite{iwen2016distributed} was proven based on the equations \eqref{AA} and \eqref{PP} to compute singular values and left singular vectors of the matrix $\mat{A}$ by finding the SVD of the proxy matrix $\mat{P}$. 
But the rank problem is arising from here, if the rank ($d$) of the block matrices of input matrix $\mat{A}$ is smaller than the rank of input matrix $\mat{A}$ itself, the algorithm could not compute singular values and singular left vectors with high accuracy. Some rows of some block matrices of $\mat{A}$ can be completely zero because of the sparsity of $\mat{A}$, so the rank of block matrices becomes smaller than $d$.
\begin{figure*}[t]
\begin{center}
\includegraphics[height=12cm,width=1.0\textwidth]{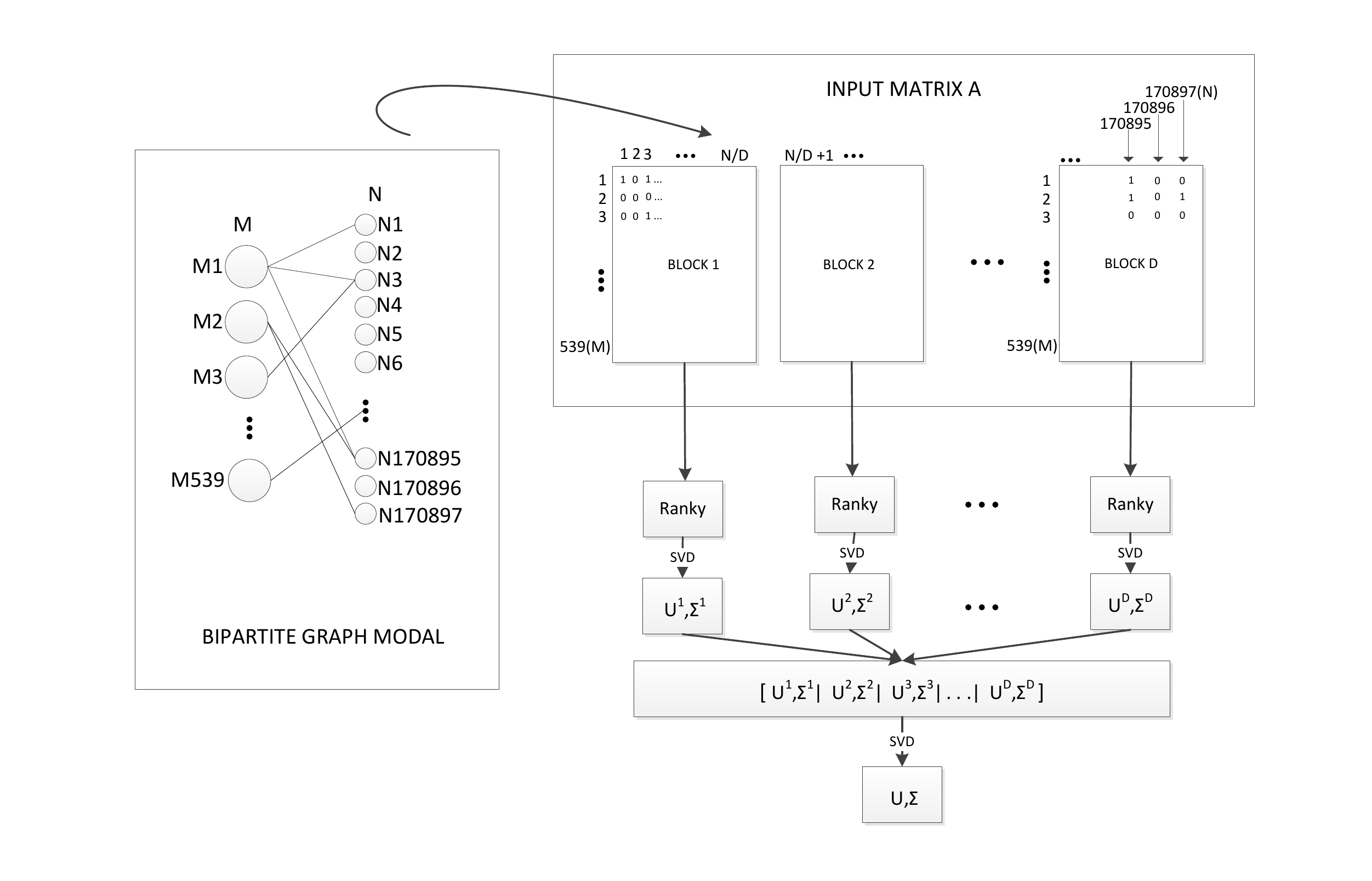}
\caption{General Schema, One Level Distributed SVD with Ranky Approach.}
\label{dist-SVD-with-ranky}
\end{center}
\end{figure*}

This problem is solved by using Ranky methods to ensure that the rank of block matrices is equal to the rank of input matrix itself. All of the process can be seen in general schema in Figure \ref{dist-SVD-with-ranky}.

RandomChecker, NeighborChecker and NeighborRandomChecker are the methods we propose to solve rank problem on large and sparse input matrix. These three methods are applied before calculating SVD of each block matrix of the input matrix. Sometimes rows of the input matrix are referred as nodes to be more descriptive. Further, the row which has no entry or contains zero in a block matrix will be called lonely node, for instance second node is a lonely node as shown in block matrix 1 in Figure \ref{dist-SVD-with-ranky}.

\iffalse
\textbf{Ranky}: The steps of Ranky are described as follows:
\begin{enumerate}
\item Load input matrix $\mat{A}$
\item Call one of the RandomChecker,NeighbourChecker or neighbourRandomChecker methods.
\item Compute singular values and singular left vectors in parallel for each block matrix.
\item Generate proxy matrix $\mat{P}$ by getting singular values and singular left vectors from all block matrices.
\item Compute singular values and singular left vectors of proxy matrix $\mat{P}$.
\end{enumerate}
\fi

\begin{algorithm}
\caption{Ranky Algorithm}
\begin{algorithmic} 
\REQUIRE input matrix $\mat{A}$  having dimension $M$x$N$
\STATE Split matrix $\mat{A}$ into D blocks based on column-wise
\FOR{$d$ = 0 \TO $D$}
  \FOR{$m$ = 0 \TO $M$}
	    \STATE $Checker = true$
        \FOR{$n$ = ($N/D)*d$ \TO $(N/D)*(d+1)$}
  	  		\IF{$A_{m,n} \neq 0 $}
	    		\STATE $Checker = false$
                \STATE break
   	 		\ENDIF
  		\ENDFOR
          \IF{$Checker$}
           \STATE call one of the RandomChecker,NeighbourChecker or neighbourRandomChecker methods
   	 	  \ENDIF
	\ENDFOR
\ENDFOR
\STATE Compute singular values and singular left vectors in parallel for each block matrix.
\STATE Generate proxy matrix $\mat{P}$ by getting singular values and singular left vectors from all block matrices.
\STATE Compute singular values and singular left vectors of proxy matrix $\mat{P}$.
\end{algorithmic}
\end{algorithm}

\begin{algorithm}
\caption{RandomChecker Method}
\begin{algorithmic} 
\STATE col = find a random column in block $d$
\STATE $A_{m,col} = 1$   
\end{algorithmic}
\end{algorithm}

\begin{algorithm}
\caption{NeighborChecker Method}
\begin{algorithmic} 
          \STATE create an empty list neighbourCandidateList          
            \FOR{$d1$ = 0 \TO $D$}
             \IF{$d1$ == $d$}
                 \STATE continue
             \ENDIF            
              \FOR{$n1$ = ($N/D)*d1$  \TO $(N/D)*(d1+1)$}
                 \IF{$A_{m,n1} \neq 0 $}
                    \FOR{$m1$ = 0 \TO in $M$}
						\IF{$A_{m1,n1} \neq 0 $}
                          \STATE add $m1$ to neighbourCandidateList
                 		\ENDIF
                    \ENDFOR
   	 		     \ENDIF
  			  \ENDFOR
  			\ENDFOR
             \STATE create an empty list neighbourList   
             \FOR{$m2$ = 0 \TO size(neighbourCandidateList)}
             	\FOR{$n2$ = 0 \TO in $(N/D)*(d+1)$}
             			\IF{$A_{m2,n2} \neq 0 $}
                          \STATE add $n2$ to neighbourList
                 		\ENDIF
                \ENDFOR
             \ENDFOR
			\STATE col = choose a random column from neighbourList
			\STATE $A_{m,col} = 1$   

\end{algorithmic}
\end{algorithm}

\begin{algorithm}
\caption{neighbourRandomChecker Method}
\begin{algorithmic} 
          \STATE Firstly call NeighborChecker method
          \STATE Then call RandomChecker method
\end{algorithmic}
\end{algorithm}

It is assumed that the number of rows is smaller than the number of columns in each block matrix. Then, the rank of each block matrix is equal to the rank of input matrix $\mat{A}$ with the approximate probability formula as below:
\begin{equation}\label{pr}
Pr \cong \left( 1 - \frac{1}{NC}*NO  \right)
\end{equation}

In Equation \ref{pr}. $NC$ represents the number of columns in the block matrix and $NO$ represents the number of rows which has only one column filled. Assume that the following block matrix having dimension of $5$ x $500$ and only the last row has no entry in any column.
\[
\begin{bmatrix}
     0 & 1 & 0 & 1 & \dots & 0 \\
     1 & 0 & 0 & 0 & \dots & 0 \\
     0 & 0 & 1 & 0 & \dots & 0 \\
     0 & 0 & 0 & 1 & \dots & 0 \\
     0 & 0 & 0 & 0 & \dots & 0 \\  
\end{bmatrix}
\]
The second, third and forth rows have only one entry in the first, third and forth column respectively. Hence number of columns for this block matrix is $NC=$500 and number of rows which has only one entry is $NO$ = 3. If the last row is filled randomly with RandomChecker method, the approximate probability of getting same rank with input matrix in terms of row-wise will be as follows:

\begin{equation}\label{Prc}
  Pr \cong \left( 1 - \frac{1}{500}*3 \right) = 0.994 = 99.4\%
\end{equation}

As shown in Equation \ref{Prc}, the approximate probability of input matrix and block matrices is $99.4\%$ when using RandomChecker method in the previous example. If the number of columns gets bigger, then the approximate probability will be higher. Although the RandomChecker method has a high approximate probability, it does not consider the neighborhood information of nodes in the graph.

The term "approximate" is used with the probability, because there is no way to calculate certain probability of matrix having rank $d$ as far as we know.
We propose another method called NeighborChecker which considers checking neighbors of each node.
\iffalse
\textbf{NeighborChecker}: This method is used just in step 3 instead of RandomChecker.
\begin{itemize}
\item Check each node of each block matrix, if there is a lonely node, then continue, otherwise go to the next step
\item Check other block matrices to determine neighbors of the corresponding node.
\item Add the candidate neighbors found in other block matrices to NeighborCandidateList
\item Find all nonzero columns of the nodes in the NeighborCandidateList and add these columns to NeighborList
\item Select a column from the NeighborList list and add an edge between the lonely node and the column in the corresponding block matrix.
\end{itemize}
\fi

Each block matrix is checked by NeighborChecker method as shown in Figure \ref{dist-SVD-with-ranky}. For instance, there are edges between M1 and N1,N3 and N170895. However, there are no edges between M2 and others in the N side in the first block matrix. But there are edges between M2 and N170895,N170897. 
NeighbourChecker is first checking the first block matrix and recognizes that the second row (M2) is completely zero, meaning that there is no edge between M2 and any nodes in the N side. Then other blocks are being checked one by one to determine neighbors of M2 and if there is a neighbor, this neighbor is added to the neighborList. In the given example, M1 is one of the neighbor of M2 because of the neighbourhood of N170895. Then a common edge between M1 and M2 is put to the second row of the first block matrix to equalize the rank of the this block matrix with the rank of the input matrix.
But this method has some disadvantages when adding an edge to lonely node. For instance, if there is only one neighbor which has only one column filled (entry) of lonely node, choosing that column causes smaller rank than $d$. Even if there are more neighbors, choosing a node which has one column filled randomly causes same problem. Therefore, neighbourRandomChecker which is a combination of first two methods can be used.

\section{Experimental Results}
\label{sec:ER}
The data used in the experiments was provided from one of the most popular online job site company in Turkey
and it consists of totally 539 nodes in one side and 170897 nodes in the other side. Input matrix is actually a bipartite graph where rows ($M$) and columns ($N$) are two disjoint sets, such that every edge either connects a vertex from $M$ to $N$ or a vertex from $N$ to $M$. Also there is no edge that connects vertices of same set. This bipartite graph is then represented it as a job-candidate input matrix whose rows correspond to jobs and columns to candidates.
%SGO: gerçek data kullandığını söyledikten sonra bunun ne datası olduğunu nasıl bipartite graph olarak temsil ettiğini anlatabilirsin.
%R : Anlattım hocam.
The input matrix $\mat{A}$ having dimension $539$x$170897$ is created using this data as shown in Figure \ref{dist-SVD-with-ranky}. 

Distributed parallel SVD algorithm is coded to find singular values and singular left vectors in a distributed and parallel manner. LAPACK SVD algorithm, dgesvd function, is used to find Singular components of each block matrix implemented in the threaded Intel MKL library. The code was written in C and run on a core i7 (8 core) machine running Linux. This algorithm currently runs on one machine but can run on distributed machines in a cluster and transfer data between the machines via sockets. Execution times are not reported in this paper as these are ultimately dependent on the number of processors and number of machines used in a distributed situation. Sum of total error is used as an evaluation metric between the true singular values($\sigma_i$) and obtained singular values(($\hat{\sigma_i}$)) using the RandomChecker, neighbourChecker and neighbourRandomChecker. Similarly, the same evaluation metric is used for the singular left vectors between true($u_i$) and obtained($\hat{u}_i$) as shown below.
$$
  e_\sigma =  \sum_{i = 1}^N |\hat{\sigma_i} - \sigma_i|
    \quad\text{and}\quad 
  e_u = \sum_{i = 1}^N |\hat{e}_i - e_i|
$$

\begin{table}[ht]
\renewcommand{\arraystretch}{1.3}
\caption{Random Checker}
\label{table_randomChecker}
\centering
\begin{adjustbox}{max width=\textwidth}
\begin{tabular}{|c||c|c|c|}
\hline
\# Blocks&Block Size& $e_\sigma$&$e_u$\\\hline	
2 & 539 x 85448  & $2.502443e-13$ & $4.052329e-10$\\\hline
3 & 539 x 56965  & $2.067235e-13$ & $3.030222e-10$\\\hline	
4 & 539 x 42724  & $3.258505e-14$ & $6.044171e-10$\\\hline
8 & 539 x 21362  & $4.130030e-14$ & $1.867252e-10$\\\hline
10 & 539 x 17089  & $4.263256e-13$ & $4.604847e-10$\\\hline	
16 & 539 x 10681  & $4.501954e-14$ & $6.100364e-10$\\\hline	
32 & 539 x 5340  & $2.554623e-13$ & $9.281878e-10$\\\hline	
64 & 539 x 2670  &$8.620882e-14$&$3.095248e-10$\\\hline	
128 & 539 x 1335  &$3.600453e-13$&$1.665984e-10$\\\hline	
\hline
\end{tabular}
\end{adjustbox}
\end{table}

\begin{table}[ht]
\renewcommand{\arraystretch}{1.3}
\caption{neighbour Checker}
\label{table_neighbourChecker}
\centering
\begin{adjustbox}{max width=\textwidth}
\begin{tabular}{|c||c|c|c|}
\hline
\# Blocks&Block Size& $e_\sigma$&$e_u$\\\hline	
2 & 539 x 85448  & $2.522729e-14$ & $1.502954e-01$\\\hline	
3 & 539 x 56965  & $4.903300e-13$ & $1.069363e-02$\\\hline	
4 & 539 x 42724  & $2.416956e-13$ & $4.489185e-01$\\\hline
8 & 539 x 21362  & $3.885781e-14$ & $4.402455e-01$\\\hline
10 & 539 x 17089  & $3.480549e-13$ & $5.048745e-01$\\\hline	
16 & 539 x 10681  & $2.601808e-13$ & $2.820104e-02$\\\hline	
32 & 539 x 5340  & $3.574918e-14$ & $6.011384e-01$\\\hline	
64 & 539 x 2670  &$2.621237e-13$ & $4.517198e-01$\\\hline
128 & 539 x 1335  &$1.404987e-13$ & $7.113150e-10$\\\hline
\hline
\end{tabular}
\end{adjustbox}
\end{table}

Table \ref{table_randomChecker} and Table \ref{table_neighbourChecker} show the results of the sum of total error for both singular values and singular left vectors when using RandomChecker and neighbourChecker methods respectively as rank controller. The results of errors are relatively small with respect to true singular values and singular left vectors in the two tables. However, it is concluded that there is no relation between the errors of singular values and singular left vectors and number of blocks due to randomness. Although it is an advantage to use high number of blocks because of speed of execution time, it can be a problem in terms of rank of the blocks. Because the more the matrix is divided, the less rank is obtained. So the RandomChecker method might change the structure of initial matrix when using more blocks because of its nature. Other method, neighbourChecker, takes the advantage of neighborhood of the node when solving the rank problem. Even using more block matrices, the structure of initial matrix will not change. However, this method does not calculate singular values and singular left vectors with small error as much as RandomChecker method does. Because lonely nodes cause having smaller rank of the block matrices than the rank of input matrix. Hence RandomChecker and neighbourChecker methods are used together as neighbourRandomChecker to solve this problem.

\begin{table}[ht]
\renewcommand{\arraystretch}{1.3}
\caption{neighbourRandom Checker}
\label{table_neighbourrandomChecker}
\centering
\begin{adjustbox}{max width=\textwidth}
\begin{tabular}{|c||c|c|c|}
\hline
\# Blocks&Block Size& $e_\sigma$&$e_u$\\\hline	
2 & 539 x 85448  & $2.298162e-14$ & $6.175930e-10$\\\hline
3 & 539 x 56965  & $1.432188e-13$ & $7.913495e-10$\\\hline	
4 & 539 x 42724  & $2.468581e-13$ & $6.211098e-10$\\\hline
8 & 539 x 21362  & $2.033373e-13$ & $8.652412e-11$\\\hline
10 & 539 x 17089  & $1.565414e-14$ & $1.504255e-10$\\\hline	
16 & 539 x 10681  & $9.953149e-14$ & $1.138005e-10$\\\hline	
32 & 539 x 5340  & $2.702838e-13$ & $4.859414e-10$\\\hline	
64 & 539 x 2670  &$1.625922e-13$&$1.827257e-10$\\\hline	
128 & 539 x 1335  &$1.404987e-13$&$7.113150e-10$\\\hline	
\hline
\end{tabular}
\end{adjustbox}
\end{table}

Table \ref{table_neighbourrandomChecker} shows the result of neighbourRandomChecker method. Similar results are obtained with the RandomChecker method. There is no difference calculating singular values and left singular vectors in terms of sum of total error when using three different methods as shown in Table \ref{table_randomChecker} and \ref{table_neighbourrandomChecker}. However, it might be better to use neighbourRandomChecker method especially for clustering problems. Graph clustering approaches aim at finding groups of densely connected nodes. Since  neighbourRandomChecker method takes the advantage of neighborliness of the nodes using this method provides connected  nodes to be in the same cluster.

\section{Conclusion and Future Work}
\label{sec:FW}
%SGO: ilk cümledüzelsin. for large sparse matrix için ne algoritması. rectangular matrix kullnamayalım diye konuşmamış mıydık?
%R : Düzelttim hocam.
This paper proposes a set of methods, called Ranky, to get singular values and left singular vectors of a large, sparse, short and fat matrix in a distributed manner. Ranky is inspired by (one-level) distributed parallel SVD algorithm. It is used before the distributed algorithm to be able to make rank of the matrix equal the rank of its block matrices. The experimental results show that Ranky algorithm is proven to recover singular values and left singular vectors of large and sparse input matrices with relatively small error. RandomChecker employs random strategy to overcome rank problem of sparse input matrix and neighbourChecker uses neighbours of the nodes. Lastly neighbourRandomChecker solves the problem by taking advantage of both methods. As for future work, some methods can be added to completely guarantee that the rank of the block matrices are equal to the rank of input matrix. Furthermore, new algorithms can be developed to get right singular vectors without distributing calculated singular values and left vectors from server to clients. 
%SGO: başlık Conclusion and Future Work ama future work yazmamışsın.
%R : Future work'u kaldırdım hocam direkt.


\section*{Acknowledgment}
We thank the www.kariyer.net for providing us the real dataset. This work was supported by I.T.U BAP project MYL-2017-40719 and the Scientific and Technological Research Council of Turkey (TUBITAK) 1505 project 5170032.
%SGO: proje numaralari yazilsin. Tubitak projesini de yazalim.



\bibliographystyle{IEEEtran}
\bibliography{bare_conf.bib}

%\begin{thebibliography}{1}


%\end{thebibliography}

\end{document}